\documentclass[conference,letterpaper,10pt]{ieeeconf}
\usepackage[left=1.91cm,right=1.91cm,top=1.91cm,bottom=1.91cm]{geometry}

\usepackage{graphicx}
\usepackage{amssymb}
\usepackage{amsmath}
\usepackage{cite}
\usepackage{comment}
\usepackage{color}
\usepackage{url}
\usepackage{alltt}
\usepackage{cleveref}
\usepackage{tikz}
\usepackage{pgfplots}
\usepackage{pgf-umlsd}
\usepackage{cprotect}
\usepackage{multirow}
\usepackage{algorithm}
\usepackage[noend]{algpseudocode}
\usepackage{bigfoot}
\usepackage{paralist}
\usepackage{tabularx}
\usepackage[whole]{bxcjkjatype}
\usepackage{ifthen}
\usepackage{threeparttable}
\usepackage[hang,small,bf]{caption}
\usepackage[subrefformat=parens]{subcaption}
\usepackage[super]{nth}
\usepackage{colortbl}
\usepackage{scalefnt}
\usepackage{threeparttable}
\usepackage{rotating}
\usepackage{caption}

\pgfplotsset{compat=1.14}
\IEEEoverridecommandlockouts

\makeatletter
\newcommand{\figcaption}[1]{\def\@captype{figure}\caption{#1}}
\newcommand{\tblcaption}[1]{\def\@captype{table}\caption{#1}}
\makeatother

\title{\LARGE \bf
Precise Well-plate Placing Utilizing Contact During Sliding\\
with Tactile-based Pose Estimation for Laboratory Automation
}

\author{
Sameer Pai$^{1*}$,
Kuniyuki Takahashi$^{2*}$,
Shimpei Masuda$^{2}$,
Naoki Fukaya$^{2}$,\\
Koki Yamane$^{3}$,
Avinash Ummadisingu$^{2}$
\thanks{$^{*}$ The starred authors have contributed equally.}
\thanks{
        $^{1}$S. Pai is with the Massachusetts Institute of Technology. This work was done during an internship at Preferred Networks, Inc.
        {\tt\footnotesize 
        sampai@mit.edu }
        $^{2}$K. Takahashi, N. Fukaya, S. Masuda, and A. Ummadisingu are with Preferred Networks, Inc. 
        {\tt\footnotesize 
        \{{takahashi, fukaya, masuda, ummavi\}@preferred.jp}}
        $^{3}$K. Yamane is with the University of Tsukuba. This work is done during a part-time job at Preferred Networks, Inc.
        {\tt\footnotesize 
        yamane.koki.td@alumni.tsukuba.ac.jp}
}
}

\begin{document}

\maketitle
\thispagestyle{empty}

\begin{abstract}
Micro well-plates are an apparatus commonly used in chemical and biological experiments that are a few centimeters thick and contain wells or divets.
In this paper, we aim to solve the task of placing the well-plate onto a well-plate holder (referred to as \textit{holder}).
This task is challenging due to the holder's raised grooves being a few millimeters in height, with a clearance of less than 1~$\mathrm{mm}$ between the well-plate and holder, thus requiring precise control during placing.
Our placing task has the following challenges:
1) The holder's detected pose is uncertain;
2) the required accuracy is at the millimeter to sub-millimeter level due to the raised groove's shallow height and small clearance;
3) the holder is not fixed to a desk and is susceptible to movement from external forces.
To address these challenges, we developed methods including a) using tactile sensors for accurate pose estimation of the grasped well-plate to handle issue (1);
b) sliding the well-plate onto the target holder while maintaining contact with the holder's groove and estimating its orientation for accurate alignment.
This allows for high precision control (addressing issue (2)) and prevents displacement of the holder during placement (addressing issue (3)).
We demonstrate a high success rate for the well-plate placing task, even under noisy observation of the holder's pose.
\footnote[4]{An accompanying video is available at the following link:\\ \url{https://www.youtube.com/watch?v=noyxIYfVxq0}}
\end{abstract}
\section{Introduction}
\label{sec:introduction}
Manipulation with precision on the millimeter to sub-millimeter order is often required when placing a grasped object onto a machine or measuring apparatus.
In factories, robots are able to achieve this precision manipulation with high speed and efficiency through extremely accurate knowledge of the pose of both the object and apparatus.
Similarly, existing laboratory automation in chemical and biological experiments typically achieves precision manipulation by keeping the apparatus fixed in place~\cite{fleischer2016application, yachie2017robotic, burger2020mobile, lim2020development, shiri2021automated, kanda2022robotic}.
However, when the experimental process or the types and positions of the apparatus change from time to time, it is time and money-consuming for robot engineers to redevelop a system for each change.
A more flexible system that does not require ongoing input from robotic engineers could enable automated experiments that are frequently changing or are not performed frequently enough to warrant permanent fixtures.

\begin{figure}[tb]
    \centering
    \includegraphics[width=0.80\columnwidth]{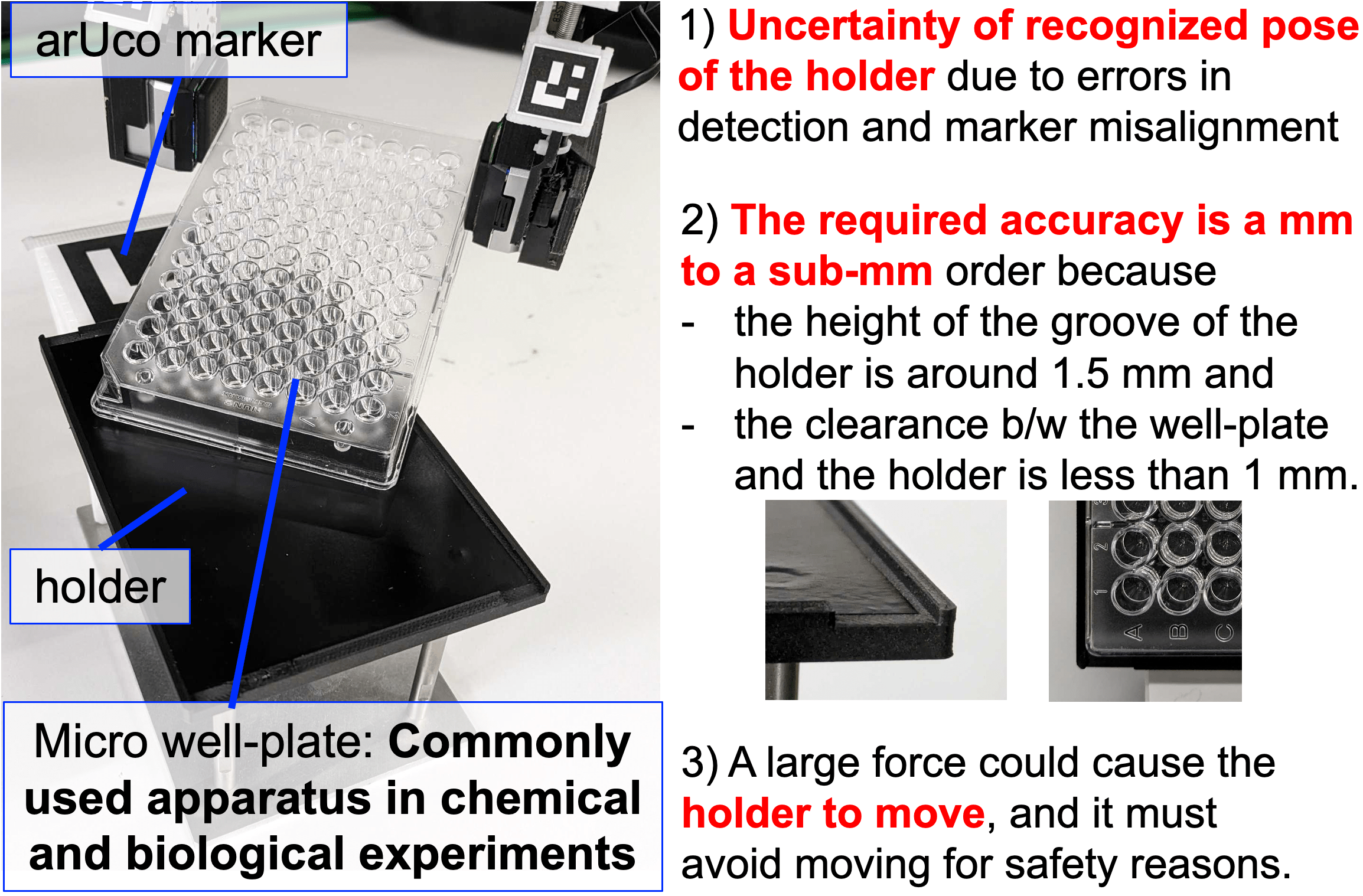}
    \caption{Challenges of placing the well-plate onto the holder}
    \label{fig:peg_insertion}
\end{figure}

One approach to achieving precise object placing is to accurately estimate the object pose (position and posture).
Achieving the necessary estimation accuracy of a millimeter to sub-millimeter order for successful placing remains a challenge, even with state-of-the-art object pose estimation methods.
Despite the capability of detecting object pose using markers, such as arUco markers~\cite{garrido2014automatic}, recognition errors and marker misalignment often lead to inaccurate detection.
To overcome the limitations in achieving precise object placing with camera image processing alone, researchers have begun to explore approaches that combine the use of force and tactile sensors~\cite{morgan2023towards,okumura2022tactile, bauza2022tac2pose, ota2023tactile}.

Our goal is to place a plate of a few centimeters in thickness, called a \textit{well-plate}, commonly used in chemical and biological experiments, onto a well-plate holder (From this point forward, referred to as the \textit{holder}).
This holder is similar to common apparatus loading areas, like those found in microplate readers or reagent dispensers.
The well-plate and holder can be located arbitrarily, providing flexibility and adaptability to various experimental setups.
To perform this task, the following issues must be addressed (Fig.~\ref{fig:peg_insertion}):
\begin{enumerate}
    \item Although the holders's poses can be detected with markers, errors in detection and marker misalignment can cause shifts of more than a few millimeters.
    \item The required accuracy is a millimeter to a sub-millimeter order because the height of the raised groove is a few millimeters and clearance between the well-plate and the holder is less than 1~$\mathrm{\,mm}$.
    \item The holder, lightweight and not fixed to a desk, can move under even minimal external force, risking collisions that may cause breakage or chemical spills.
\end{enumerate}

To address these issues, we developed a well-plate placement method with two key characteristics:
\begin{itemize}
    \item We address the issue (1) through the accurate estimation of the well-plate pose using a tactile sensor after the gripper grasps it.
    This allows us to minimize the impact of potential errors in marker detection.
    \item To address issues (2) and (3), we use a precise placing method in which the well-plate is slid onto the holder while maintaining lateral contact with the holder's raised groove, aligned with its estimated orientation.
    This method ensures alignment despite the limited clearance and prevents the holder from moving, keeping it stationary throughout the placement process.
\end{itemize}

\section{Related Works}
\label{sec:related_works}
Our task involves placing the well-plate onto the holder.
Since the holder is surrounded by the raised grooves, it can also be perceived as a `hole.'
Therefore, the task can also be interpreted as a peg-in-hole task, where the well-plate is inserted into this shallow `hole.'
This section discusses research related to these two perspectives: placing and insertion.

\subsection{Related Works on Object Placing Task}
\label{sec:related_works_placing}
A typical method for placing an object would be to estimate the object's pose.
Many image-based studies exist for object pose estimation, but these methods are known to encounter occlusion problems and have limited precision~\cite{sun2022onepose, Li2023vox}.
Note that another method involves positioning the grasped object in contact with the ground or other objects to reduce positional uncertainty and enable more accurate pose estimation~\cite{von2020contact, pankert2023learning}.
These methods require prior knowledge of the object's shape and pose, particularly of the parts that make contact.
Moreover, they can be challenging to apply when there is ambiguity in the object's placement pose.

Studies using tactile sensors have demonstrated high accuracy in pose estimation, mainly due to their ability to capture fine-grained contact information~\cite{anzai2020deep, lach2023placing}.
In particular, vision-based tactile sensors, such as GelSight, have been employed in research, with some even achieving millimeter to sub-millimeter accuracy~\cite{anzai2020deep, okumura2022tactile}.
Therefore, in this study, tactile sensors are utilized for object pose estimation.

While most tactile studies don't specify the placement location, assuming a flat desk surface, our study requires accurate object placing within raised grooves with under 1~$\mathrm{\,mm}$ clearance.
Moreover, the placement location information can suffer from a few millimeters of noise due to errors in marker recognition.

\subsection{Related Works on Object Insertion Task}
\label{sec:related_works_insertion}
The object insertion task has been studied from the perspectives of both hardware-based and software-based approaches.
The hardware approach exploits compliance to absorb uncertainties in the object's shape and pose, which prevents any potential breakage of the object.
This compliant behavior can prove beneficial in tasks demanding millimeter to sub-millimeter precision, even without rigid control~\cite{wang2019robotic, von2020compact}.
Often in these approaches, it's assumed that the insertion task is executed by pressing a compliance mechanism against the target of insertion, which necessitates that the target is fixed and does not move during the process.
Consequently, when the weight of the insertion target is light, as in our task, and can be moved by external forces, it becomes challenging to apply these methods.
Note that although our study uses an adaptive finger with compliance, it is only used to prevent damage to the tactile sensor and does not realize insertion by hardware.
We will show in \cref{sec:result_peg_insertion} that while our adaptive finger does indeed prevent the damage of the tactile sensor, it does not contribute to the success of our placing task.

\begin{figure}[t]
    \centering
    \includegraphics[width=0.90\columnwidth]{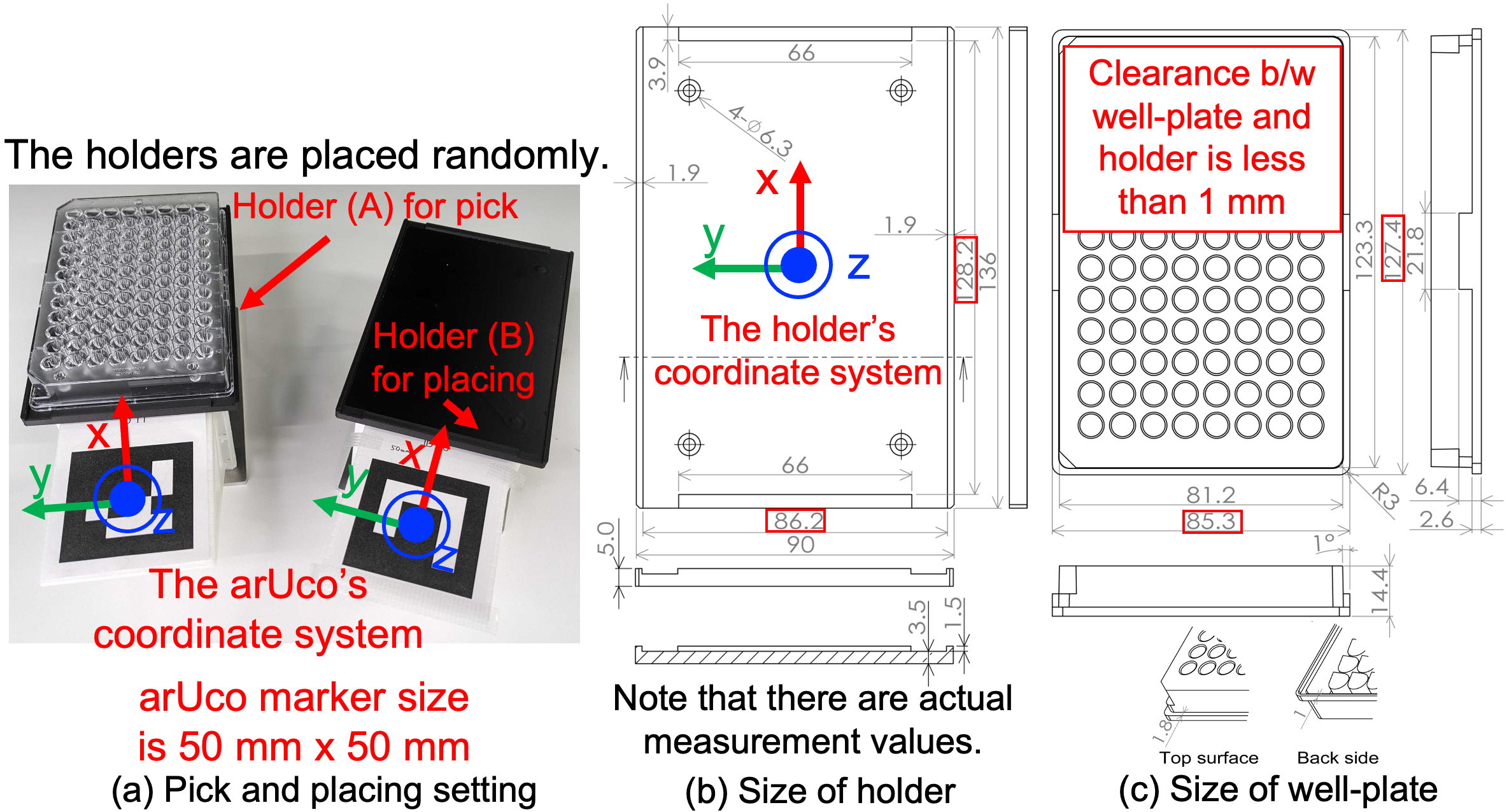}
    \caption{Task setup and size of the holder and well-plate.}
    \label{fig:well_plate}
\end{figure}

In our task setting, the height of the raised grooves of the holder is 1-2~$\mathrm{\,mm}$, and the width is approximately 2~$\mathrm{\,mm}$.
Additionally, the location to place the well-plate on the holder is not completely encircled by these raised grooves; there are gaps (Fig.~\ref{fig:peg_insertion} \& Fig.~\ref{fig:well_plate}).
Consequently, there's a possibility that the well-plate could be placed in a manner where it either overshoots or rests on the raised groove instead of fitting precisely into the holder's cavity.
However, such placing would be considered a failure.

Software-based approaches to insertion tasks primarily involve the insertion of a peg into a hole.
However, these scenarios often involve deep holes situated on a flat surface, differing greatly from our task setup.
Applying techniques from previous studies to our context, hence, is challenging. 
A common approach in existing research utilizes Force/Torque (F/T) sensors or tactile sensors to estimate the location of a hole~\cite{fei2003assembly, morgan2023towards}.
There are achieved by moving the peg across the surface before inserting it into the hole.
However, using this method in our task could result in the well-plate not fitting snugly into the designated location but overshooting the raised groove due to the wide of the groove being small.
Another method estimates contact points and edges by pivoting motion of a peg in the process of peg-insertion~\cite{kim2022active}.
However, this approach does not work well because the plate and the surface of the holder come into contact because the height of the groove of the holder is shallow.

Given these challenges with existing methodologies, we propose a new method for the accurate placing of a well-plate.
The approach entails maintaining contact between the well-plate and the raised groove of the holder while executing a sliding motion to better ensure accurate placement.

\begin{figure*}[t]
    \centering
    \includegraphics[width=1.70\columnwidth]{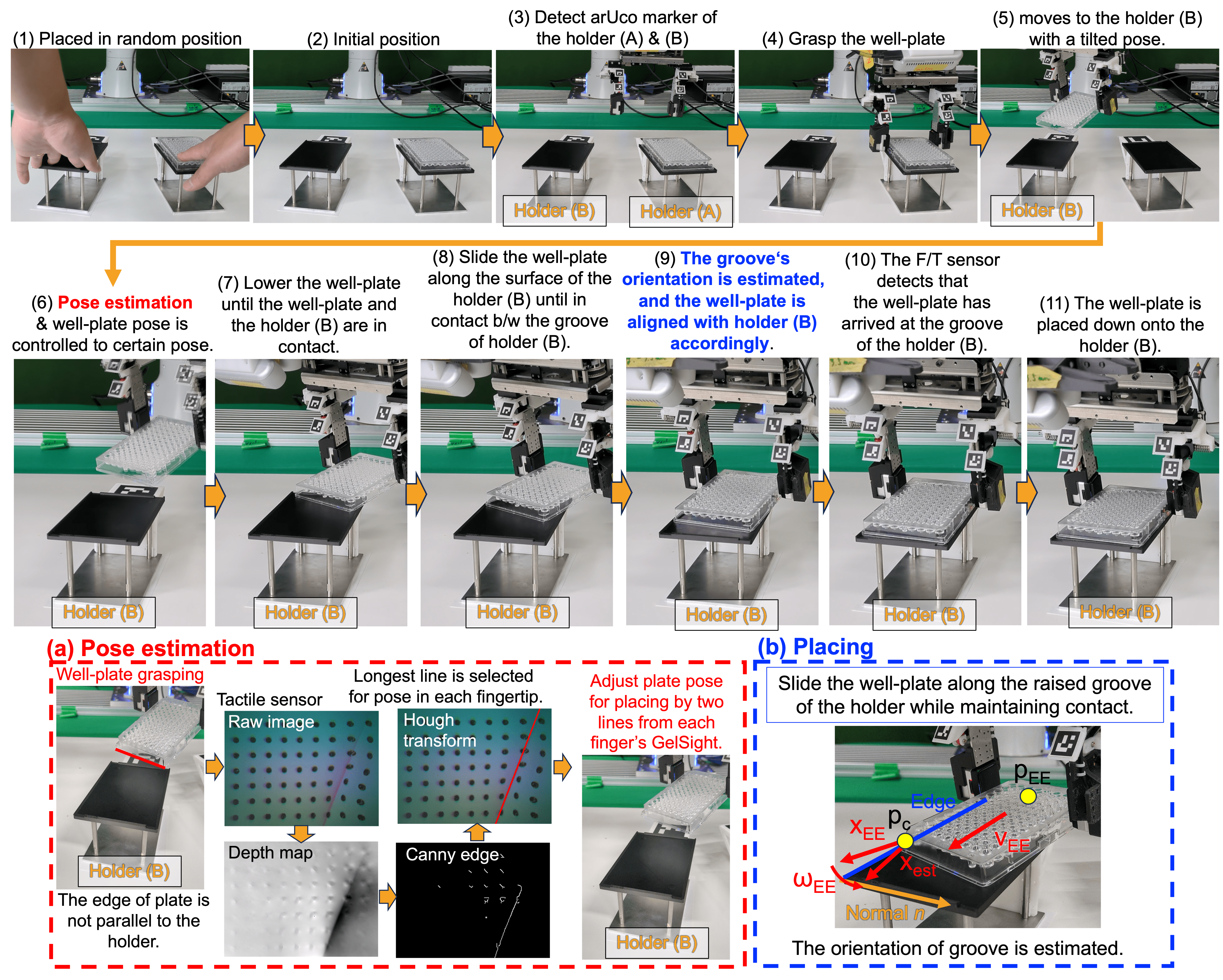}
    \caption{Overview of pick and place, and two key components of our proposed method: pose estimation and placing.}
    \label{fig:method}
    \vspace{-6mm}
\end{figure*}
\section{Task Setup}
\label{sec:Task Setup}
As depicted in Fig.~\ref{fig:well_plate}, our task setup uses two holders, each comprising two flat plates supported by four pillars.
Importantly, these holders are not affixed and are able to move when there is an appropriate application of external force, such as when the well-plate is placed onto a holder with significant force.
The designated plate on which the well-plate is placed is created with a 3D printer modeled after standard laboratory apparatus.
To compensate for any unevenness on its surface that might be caused by the 3D printer's printing accuracy, we attach a 0.1~$\mathrm{\,mm}$ thick sticker to it. 
The height of the raised grooves with the thick sticker is 1.4~$\mathrm{\,mm}$.
The size of the well-plate is $85.3~\mathrm{\,mm} \times 127.4~\mathrm{\,mm}$, and the size of the inside groove of the holder is $86.2~\mathrm{\,mm} \times 128.2~\mathrm{\,mm}$.
The clearance between the well-plate and holder is less than 1~$\mathrm{\,mm}$ (Fig.~\ref{fig:well_plate}~(c)).
For our tasks, we use an arUco marker of size 50~$\mathrm{\,mm}$, attached to a holder that is fastened onto the pillars of the holder.
We chose arUco markers in our setup over other various marker systems available.
As highlighted in \cite{kalaitzakis2020experimental}, the accuracy across different marker systems doesn't vary significantly, but we find the arUco markers score high on ease of use.
\section{Method}
\label{sec:method}
We will begin by outlining the pick-and-place process in \cref{sec:process of pick and insertion}, which provides an overview of the methodology.
Subsequently, we dive into the two key components of our method: A) Pose estimation of the well-plate using tactile sensors, detailed in \cref{sec:method_pose_estimation}, and B) Placing during sliding contact, which is achieved by accurately estimating the raised groove's orientation, described in \cref{sec:method_peg_insertion}.
\subsection{Overview of Pick and Place}
\label{sec:process of pick and insertion}
The setup involves a well-plate on holder~(A) and an empty holder~(B), each of which is marked with an arUco marker (Fig.~\ref{fig:well_plate}~(a) and  Fig.~\ref{fig:method}~(1) \& (2)).
Note that since the relative 6D pose of the holders from the arUco marker is known, the initial configurations of holder~(A) and (B) on the table do not have an effect on the grasping or insertion behavior once the arUco markers have been detected.
For the sake of experimental diversity, the holders for pick and place are initially manually arranged on the desk in random positions (Fig.~\ref{fig:method}~(1) \& (2)).

First, the robot detects the arUco markers on holders~(A) and (B) (Fig.~\ref{fig:method}~(3)).
200 images are captured, and a low-pass filter is applied to enhance the detection accuracy of the arUco marker.
Despite this, minor deviations of 1-2~$\mathrm{\,mm}$ and 1-2~degrees can still occur, arising from inherent limitations in the marker recognition precision.
Although minimal, these discrepancies in marker detection can significantly affect the precision of the well-plate placement process.

Next, the robot moves to the position of the holder~(A), grasps the well-plate (Fig.~\ref{fig:method}~(4)), and moves to a height of 4~$\mathrm{\,cm}$ from the holder~(B) with a tilted pose (Fig.~\ref{fig:method}~(5)).
The path planning for the movement of the robot from the holder~(A) to the holder~(B) was done using MoveIt!~\cite{coleman2014reducing}.
Then, using the tactile sensors, the well-plate pose is estimated, and the well-plate pose of roll, pitch, and yaw is controlled to be $(0\,deg, 15\,deg, 25\,deg)$ in the holder~(B)'s coordinate system (Fig.~\ref{fig:well_plate}~(b) and Fig.~\ref{fig:method}~(6)).
This particular orientation was determined through a process of direct teaching.
This process involves manually guiding the robot to adopt the desired pose, which is recorded and reproduced during the experiment.
This rotational maneuver notably increases the tolerance of the placing movement.
This enhancement allows for a significant level of inaccuracy in the estimation of the holder's position, thereby making the process more robust to uncertainties.

Then, under velocity control to maintain smooth movement, the well-plate is lowered towards the holder~(B).
The descent is stopped once the well-plate makes contact with the holder~(B), a state detected from observation of the well-plate pose using tactile sensors (Fig.~\ref{fig:method}~(7)).
Under velocity control, the robot slides the well-plate along the surface of the holder~(B).
When the well-plate contacts the raised groove of the holder~(B) during sliding, the robot estimates the raised groove's direction.
It then moves the well-plate in this direction, aligning it longitudinally with the holder~(B) (Fig.~\ref{fig:method}~(8)).
Sliding the well-plate along the holder~(B)'s raised groove ensures continuous contact and alignment  (Fig.~\ref{fig:method}~(9)).
It also helps prevent applying excessive force, which could lead to undesired movements of the holder~(B) and well-plate.
These actions allow for a precise and secure placement of the well-plate onto the holder~(B).
When the well-plate contacts the raised groove of the holder~(B), the F/T sensor reading increases beyond a set threshold.
This indicates that the well-plate has arrived in the correct position (Fig.~\ref{fig:method}~(10)).
The robot then places the well-plate down onto the holder~(B) (Fig.~\ref{fig:method}~(11)).

\subsection{Utilization of Tactile Sensors in Pose Estimation}
\label{sec:method_pose_estimation}
In this section, we describe the pose estimation of the well-plate using two GelSight Mini tactile sensors~\cite{gelsight2024}, each mounted on a robot fingertip.
GelSight is characterized by its camera-embedded elastomeric fingertip, which captures high-resolution RGB images of the surface that makes contact with objects. 
After a successful grasp, the contact area between the sensor and the well-plate forms a single edge (Raw image in Fig.~\ref{fig:method}~(a)).
To detect the pose of this edge, we first process the GelSight Mini RGB image into a Depth Map by deep learning, as described in~\cite{wang2021wedge}.
The Depth Map is an estimation of the depth information of an object that is pressed against the GelSight, derived from the raw RGB image using a neural network provided from~\cite{gelsight2024}.
The use of a Depth Map is necessary because the surface of our GelSight is drawn with black dots.
When the well-plate is pressed against the tactile sensor, the edge of the well-plate is visible in the raw RGB image, but it appears much weaker compared to the black dots.
Even after grayscaling and edge extraction, only the black dots of GelSight are extracted, preventing the successful extraction of the well-plate edges.
A comparison of the results with and without Depth Map (on raw RGB) will be described in \cref{sec:results_edge_detection}.

We then take the Depth Map and use the Canny edge detector to find areas of high gradient in the map.
Finally, we take the output of the Canny edge detector and use the Hough transform to find a straight line in the image.
If multiple lines are generated by the Hough transform, we opt for the longest one.
This is because longer lines represent a larger area of contact between the sensor and the well-plate, which tends to be less affected by sensor noise, resulting in a more reliable estimation.

We perform this line detection on both sensors.
Each line detected corresponds to the contact edge against the well-plate.
We then find a plane that fits these two lines to estimate the pose of the well-plate.
Specifically, we calculate the axes parallel to each line (contact edge) and the sub-axes perpendicular to them.
These correspond to the long and short edges of the well-plate, respectively.
By averaging these axes and sub-axes between the two sensors, we compute the median axes, which represent the estimated pose of the well-plate.
With these median axes computed, we calculate the midpoint between the line of contact of the well-plate with the two tactile sensors to serve as the midpoint of the estimated pose.
The estimated pose provides us with both the orientation (how the well-plate is tilted or pointed) and position (where it is located) of the well-plate in relation to the robot's fingertips.
We use this information to adjust the well-plate pose for the successful placing of the well-plate.
\subsection{Placing utilizing Contact during Sliding}
\label{sec:method_peg_insertion}
In this section, we describe placing the well-plate by estimating the direction of the holder's raised groove.
This method enables the well-plate to maintain contact with the raised groove of the holder~(B), thereby avoiding the application of unnecessary external forces to the holder~(B).
Throughout this section, $p_{EE}, R_{EE}, v_{EE}, \omega_{EE}$ denote the world-frame position, rotation matrix, velocity, and angular velocity of the end-effector, respectively, as illustrated in Fig.~\ref{fig:method}~(b).
Furthermore, a point $x$ located in the end-effector frame would have its corresponding coordinates $p_{EE} + R_{EE} x$ in the world frame.

We use a \textit{velocity-based force controller} to enable commanding a desired force $F_{des}$ against the environment.
Given the desired force $F_{des}$ and the current force $F_m$ from the F/T sensor (calibrated), we impose a proportional velocity law as follows:
\begin{equation}
    v_{EE} = k_F(F_m - F_{des}), \label{eq:control_law}
\end{equation}
where $k_F$ is a control constant.
This law implies that the velocity of the end-effector is dynamically responsive to the difference between the measured and desired force.

Using this controller, we design a multi-phase placing process for the well-plate.
This process takes inspiration from the typical human approach to the task of placing a well-plate, which often involves initially positioning one edge of the well-plate and then sliding the rest into place.

First, the robot rotates the well-plate by a predefined angle and then lowers it into the holder~(B) until contact is detected (Fig.~\ref{fig:method}~(7)).
Once the edge of the well-plate makes initial contact with the surface of the holder~(B) (Fig.~\ref{fig:method}~(8)), the robot applies the desired force along the longer axis of the well-plate, systematically guiding the plate into the raised groove of the holder~(B) (Fig.~\ref{fig:method}~(9)).

Once the end-effector makes contact with the holder~(B), it becomes crucial to monitor the end-effector's motion over time.
First, we consider the movement along $\mathbf{n}$, the axis normal to the holder~(B)'s raised groove as seen in Fig.~\ref{fig:method}~(b).
On this axis, \cref{eq:control_law} gives us that 
\begin{equation}
    \mathbf{n}\cdot v_{EE} = k_F\mathbf{n}\cdot (F_m - F_{des}).
\end{equation}
This implies that if the desired force along $\mathbf{n}$ is reached, $\mathbf{n}\cdot v_{EE} = 0$, and so the end-effector moves perpendicular to the normal. 
In contrast, on the axis parallel to the holder~(B)'s geometry, the only force acting on the end-effector is static friction, meaning that $v_{EE}$ will always have a component on this axis.
We can conclude, then, that the overall trajectory of the well-plate after contact under this control law is a sliding motion along the raised groove of the holder~(B).
Since the post-contact displacement is entirely parallel to the holder~(B), we can collect the position data over time and then apply linear regression to estimate the orientation of the holder~(B)'s raised groove.

Once the holder~(B)'s raised groove orientation is accurately estimated, we continue the sliding process by orienting the end-effector to be parallel to the holder~(B)'s raised groove (Fig.~\ref{fig:method}~(9)).
We do this by applying proportional control to the end-effector's orientation.
Specifically, let $\mathbf{x}_{est}$ be the estimated orientation of the holder~(B)'s raised groove, and $\mathbf{x}_{EE}$ be the current orientation of the well-plate's edge.
Then, we command the angular velocity:
\begin{equation}
    \omega_{EE} = k_{R} (\mathbf{x}_{EE}\times \mathbf{x}_{est}) \label{eq: angular_law},
\end{equation}
where $k_{R}$ is a control constant.
However, the commanded rotation is about the center of the end-effector, and, therefore, would cause the contact point between the well-plate and the holder~(B) to move.
This movement would result in the well-plate losing contact with the holder~(B) and, therefore, a failed placing.
To fix this, we assume that there is a fixed translation $p_{c}$ from the end-effector to the contact point between the well-plate and the holder~(B) so that the contact point's position in the world frame is $p_{EE} + R_{EE} p_c$ (Fig.~\ref{fig:method}~(b)).
Then, the velocity of the contact point due to the above angular velocity is simply $\omega_{EE} \times R_{EE} p_c$.
Under this assumption, we can ensure that the contact point's velocity is zero by adding a feed-forward term to the velocity control law that counteracts this velocity:
\begin{equation}
    v_{EE} = k_F(F_m - F_{des}) - \omega_{EE}\times R_{EE} p_c. \label{eq:control_law_ff}
\end{equation}
Combining positional angular control with force-based positional control ensures smooth sliding while maintaining contact with the holder~(B).

However, the translation $p_c$ isn't known before the sliding process begins.
To estimate it during sliding, consider again the normal $\mathbf{n}$ to the holder~(B)'s raised groove.
Since the raised groove is linear, it can be modeled by the equation $\mathbf{n} \cdot \mathbf{x} = c$ for some constant $c$.
Then, at any point in time that the well-plate is in contact with the holder~(B), the contact point $p_{EE} + R_{EE} p_c$ must be on the linear boundary of the holder~(B), giving us the equation 
\begin{equation}
    \mathbf{n}\cdot(p_{EE} + R_{EE} p_c) = c.
\end{equation}
This is a linear constraint in $p_c$ and $c$, and so if we collect many samples of $p_{EE}, R_{EE}$ while contact occurs, we can again use linear regression to estimate both the holder~(B)'s location and the contact point in real-time.
Once $p_c$ is estimated, we can use the angular control with the feed-forward velocity above to complete the insertion motion.

To summarize, the placing technique consists of the following stages:
\begin{enumerate}
    \item Insert one edge of the well-plate at an angle to increase failure tolerance.
    \item Push the well-plate into the holder without rotation to regress the holder's orientation.
    \item Using the estimated orientation and end-effector position data, estimate the contact point $p_c$.
    \item Use proportional angular control to orient the well-plate while sliding and maintaining contact to complete the placing. 
\end{enumerate}
\begin{figure}[tb]
    \centering
    \includegraphics[width=0.90\columnwidth]{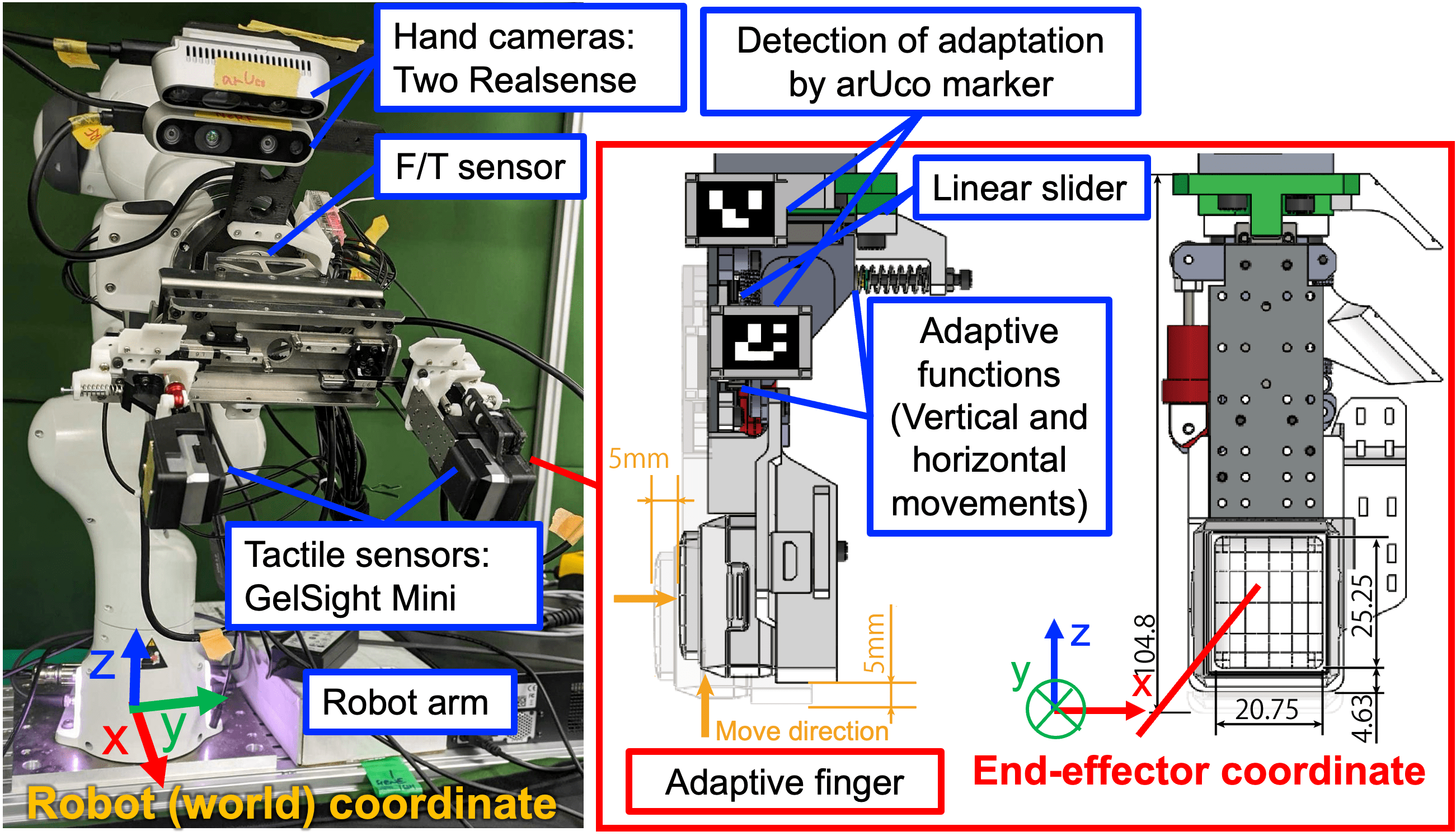}
    \caption{Robot setup used for the experiment}
    \label{fig:robot_setup}
\end{figure}
\section{Experimental Setup}
\label{sec:experimental setup}
\subsection{Robot Setup}
\label{sec:Robot Setup}
Our robotic system, shown in Fig.~\ref{fig:robot_setup}, consists of a Franka Emika Panda Arm 7-DoF robotic arm.
This arm is equipped with a custom-designed parallel gripper that serves as an end-effector, described in detail in \Cref{sec:custom_gripper}.
The gripper includes two vision-based tactile sensors, the GelSight Minis~\cite{gelsight2024}.
A Leptrino F/T sensor (FFS055YA501U6) is mounted between the robot arm and the gripper.
The rated capacity of force of F/T sensor is $F_{xyz}= \pm 500 \mathrm{\,N}$, torque is $T{xyz}= \pm 4 \mathrm{\,Nm}$, resolution is $\pm1/2000$, and sampling rate is $200 \mathrm{\,Hz}$.
In addition, we use an Intel RealSense Depth Camera D435 to oversee the workspace and an Intel RealSense Depth Camera D435i for the gripper.
The camera (D435) captures RGB images to detect arUco markers used for the pick-and-place tasks that involve positioning the well-plate.
Note that no depth information is retrieved in this study.
The controller PC runs on Ubuntu 20.04.6 LTS with ROS Noetic and is equipped with 32\,GB RAM, an Intel Core i7-7700K CPU, and a GeForce RTX 2060 6G Rev.A.

\subsection{Custom-designed Parallel Gripper}
\label{sec:custom_gripper}
Several factors, such as recognition errors and marker misalignment, may lead to discrepancies in object position estimation.
Performing grasp operations under these conditions could potentially damage the grasped object or the robot itself.
Therefore, to mitigate these errors at the hardware level and thus ensure safe pick-and-place tasks, we have developed specific features in our custom-designed parallel gripper.
Our gripper is equipped with adaptive fingers that can move in two directions, horizontally and vertically.
Fig.~\ref{fig:robot_setup} gives an overview diagram and illustrates the finger's movement.
The gripper, which is operated by a servomotor (Dynamixel XM430-W350-R), includes adaptive fingers, two GelSight Mini tactile sensors~\cite{gelsight2024}, and a pair of Intel RealSense Depth Cameras, models D435 and D435i (Fig.~\ref{fig:robot_setup}).

The adaptive finger, designed with a linear slider, allows the fingertip to smoothly slide up to 5~$\mathrm{\,mm}$.
Depending on the task, the horizontal springs—which bear spring constants of 0.067~$\mathrm{\,N/mm}$—can be easily replaced with springs of different constants.
For the vertical direction, we use springs with a constant of 0.074~$\mathrm{\,N/mm}$, which exerts a force that swiftly returns the finger to its initial position upon the object's release.

The adaptive finger moves passively when making contact with an object.
While it is often assumed that precise manipulation and object pose estimation require accurate finger positioning, this may not always be the case.
To investigate this assumption, we have equipped our system with arUco markers, allowing us to accurately measure the pose of the fingers.
The adaptive finger's displacement is gauged using two arUco markers—one attached to the base of the finger and the other to the movable part.
These markers, each of which is $15~\mathrm{\,mm}$ in size, are detected by the RealSense D435i.
In this study, we carefully position the RealSense D435i camera to minimize the chance of occlusions interfering with the detection of the arUco markers.
In the experiment section (\cref{sec:results}), we compare the results of experiments performed with and without taking the adaptive finger displacement measurements into account in order to determine the impact of this factor on the overall accuracy of pose estimation and placement.
\begin{figure}[t]
    \centering
    \includegraphics[width=0.90\columnwidth]{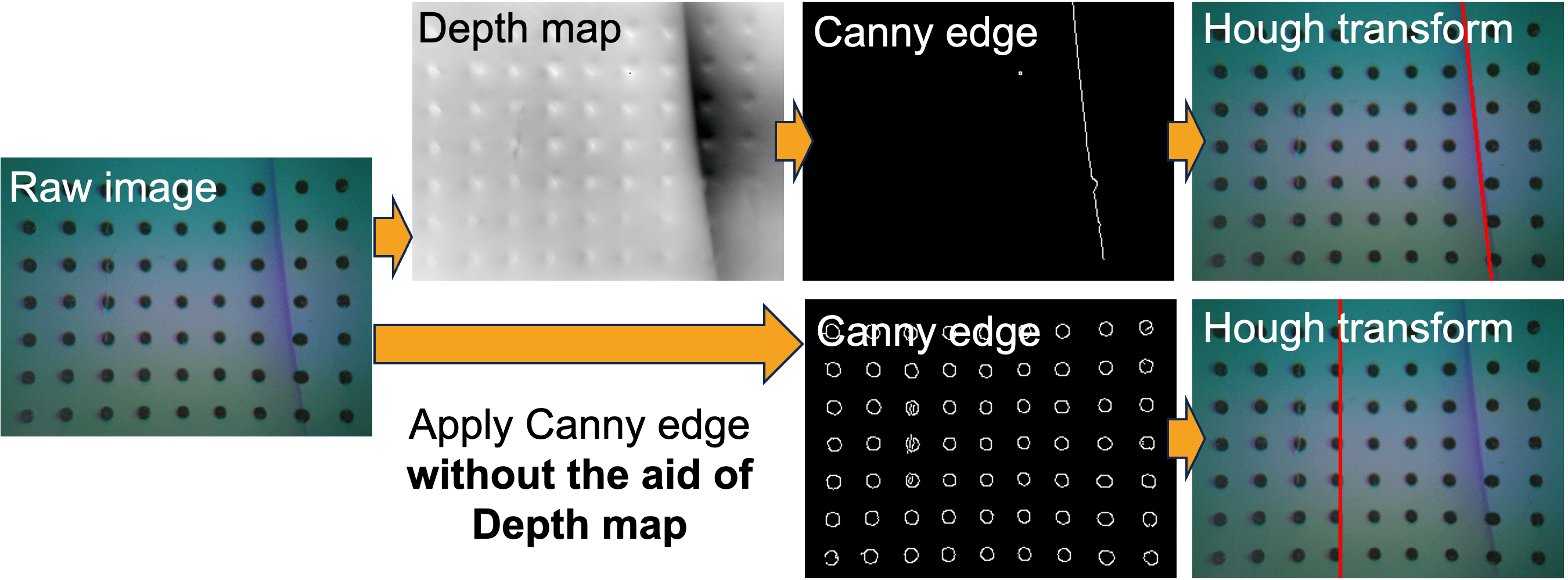}
    \caption{Comparison of the well-plate edge extraction with and without the use of Depth Map (on a Raw image).}
    \label{fig:edge_detection}
\end{figure}
\section{Experimental Results}
\label{sec:results}
In this section, we evaluate the following three items:
A) effect of using a Depth Map for the well-plate edge extraction;
B) accuracy of pose estimation using tactile sensors with adaptive fingers (\Cref{sec:results_pose_estimation});
C) evaluation of the well-plate placing task (\Cref{sec:result_peg_insertion}).
\subsection{Well-plate's Edge Detection using Tactile Sensor}
\label{sec:results_edge_detection}
We examine the effect of using a Depth Map for the well-plate edge extraction, considering scenarios both using Depth Map and using raw RGB without the aid of a Depth Map.
As shown in Fig.~\ref{fig:edge_detection} of the Raw image, when the well-plate makes contact with the tactile sensor, the edge of the well-plate appears in the raw image.
However, compared to the GelSight's black dots, the well-plate's edge appears significantly fainter.
By using a Depth Map, it facilitates an accurate recognition of the pressed well-plate's edge using the Hough Transform.
On the other hand, performing edge extraction using a Canny edge detector on the raw image, without the aid of a Depth Map, only results in the extraction of the GelSight's black dots.
This impedes the successful extraction of the well-plate edges.

\begin{figure}[t]
\begin{tabular}{cc}  
    \begin{minipage}[c]{0.35\columnwidth}
        \centering
        \includegraphics[width=0.60\columnwidth]{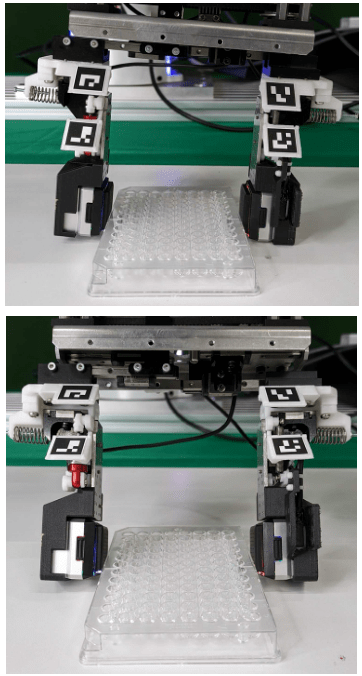}
        \figcaption{Pose examples
        }
        \label{fig:evaluation_poses}
    \end{minipage}

    \begin{minipage}[c]{0.35\columnwidth}
        \centering
        \captionsetup{width=1.5\linewidth}
        \tblcaption{The accuracy of pose estimation.}
        \begingroup
        \scalefont{0.70}
        \begin{tabular}{c||cc }
            \hline
            \shortstack{Type\\{}\\{}\\{}\\{}} & \shortstack{w/Adaptive\\Finger\\displacement\\info} & \shortstack{w/o Adaptive\\Finger\\displacement\\info}\\
            \hline\hline
            roll & $\bf{0.17\pm0.10}$ & $2.43\pm1.36$ \\
            pitch & $\bf{0.34\pm0.24}$  & $0.44\pm0.34$ \\
            roll \& pitch & $\bf{0.52\pm0.29}$  & $2.87\pm1.35$ \\
            \hline
        \end{tabular}
        \label{tab:result_pose_estimation}
        \endgroup
    \end{minipage}
\end{tabular}
\end{figure}
\subsection{Accuracy of Pose Estimation with Adaptive Finger}
\label{sec:results_pose_estimation}
We evaluate the accuracy of pose estimation, focusing on scenarios with and without considering the displacement information of the adaptive finger.
The fingertip pose is either calculated based on this displacement or kept constant.

The well-plate was positioned on a horizontal desk as shown in Fig.~\ref{fig:evaluation_poses}.
We calculate the well-plate's pose, given its position in the robot coordinate system (Fig.~\ref{fig:robot_setup}), relative to the desk's horizontal plane.
This enables us to utilize the desk's horizontal position as a ground truth reference for the well-plate's pose.
Since the parallel gripper is used to grasp a plate, the surfaces of the fingers and the object are parallel, so the yaw value is not evaluated.
We let the robot perform pose estimation of the well-plate ten times while changing the pose of the gripper relative to the plate.
The error between the estimated well-plate pose using GelSight and ground truth is calculated.
Table~\ref{tab:result_pose_estimation} shows the mean and standard deviation of the pose estimation error from 10 trials.

Both roll and pitch accuracy improved significantly when taking into account the adaptive finger displacement.
Disregarding this adaptive finger displacement information led to substantial roll errors, mainly because vertical finger displacement (z-axis in end-effector coordinate in Fig.~\ref{fig:robot_setup}), rather than horizontal (y-axis in end-effector coordinate), significantly affects the object's pose calculation.
We apply various magnitudes of external force to the adaptive finger in this experiment.
Thus, not considering the adaptive finger’s displacement increases both the error and variance in pose estimation significantly.
The error in roll pose estimation varied based on the consideration of the adaptive finger's displacement.
The influence of this error on the placing is discussed in the following section.
\begin{figure}[t]
    \centering
    \includegraphics[width=0.99\columnwidth]{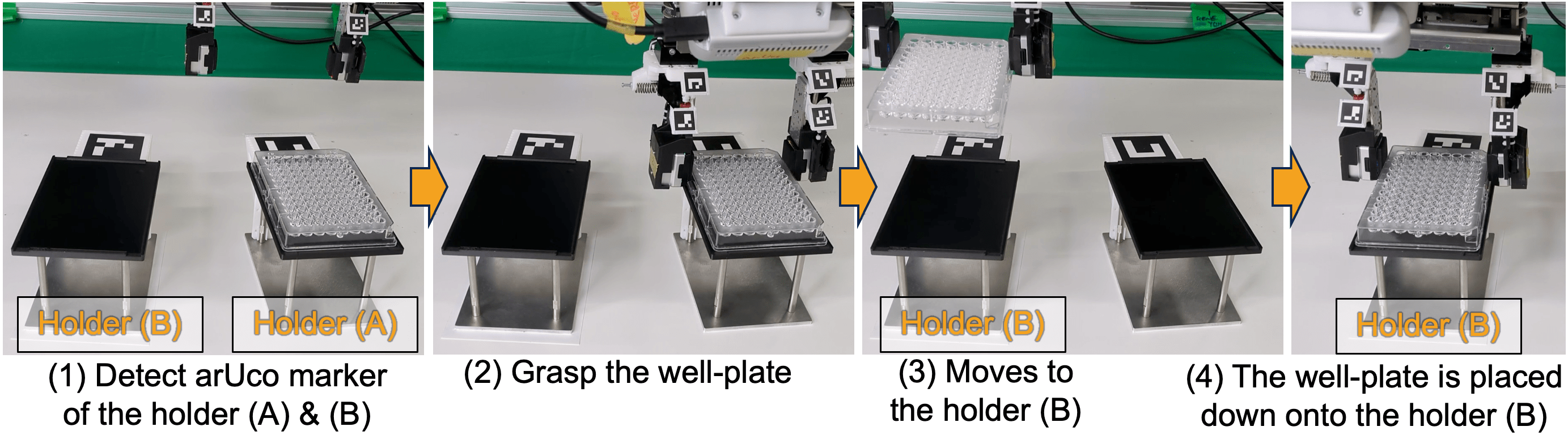}
    \caption{Simple placing (Baseline) approach}
    \label{fig:simple_placing}
\end{figure}
\begin{table*}[t]
    \centering
    \caption{Results of placing task}
    \label{tab:result_peg_insertion}
    \begingroup
    \scalefont{0.82}
    \begin{tabular}{l|cccccc||c|cc}
    
\hline
\multirow{4}{*}{Method} &
\multicolumn{6}{c||}{Conditions of method} &
\multirow{4}{*}{\shortstack{Nummer\\of\\Success}} & 
\multicolumn{2}{c}{Error size} \\
\cline{2-7}
\cline{9-10}

&
\shortstack{(1) Adaptive Finger\\displacement info\\(AFD)} &
\shortstack{(2) Pose\\estimation\\by Tactile} &
\shortstack{(3) Raised\\groove\\estimation} &
\shortstack{(4) Adaptive\\finger\\(AF)} &
\shortstack{(5) Simple\\placing} &
\shortstack{(6) Inaccurate\\the holder pose\\by noise} &
&
\shortstack{Translation\\$[mm]$}&
\shortstack{Rotation\\$[degree]$}
\\
    
\hline\hline
(a) (Ours)      &            & \checkmark & \checkmark & \checkmark  &            &            & \bf{10/10}  & $11.9\pm6.9$ & $4.5\pm1.6$ \\
(b) (Ours w/ AFD)     & \checkmark & \checkmark & \checkmark & \checkmark  &            &            & \bf{10/10}  & $10.1\pm5.7$ & $4.2\pm1.0$ \\
(c)             & \checkmark &            & \checkmark & \checkmark  &            &            &     0/10    & - & - \\
(d)             & \checkmark & \checkmark &            & \checkmark  &            &            &     0/10    & - & - \\
(e)             & \checkmark & \checkmark & \checkmark &             &            &            & Tactile broken  & - & - \\
(f) (Baseline)  &            &            &            & \checkmark  & \checkmark &            &     6/10    & - & - \\
(g) (Baseline w/o AF) &            &            &            &             & \checkmark &            &     6/10    & - & - \\
\hline
(a$^{*}$) (Ours)      &            & \checkmark & \checkmark & \checkmark  &            & \checkmark & \bf{10/10} & $10.8\pm8.9$ & $2.8\pm1.02$ \\
(b$^{*}$) (Ours w/ AFD)     & \checkmark & \checkmark & \checkmark & \checkmark  &            & \checkmark & \bf{10/10} & $12.1\pm7.6$ & $5.1\pm0.42$ \\
(f$^{*}$) (Baseline w/o AF)  &            &            &            & \checkmark  & \checkmark & \checkmark &     0/10  & - & - \\
\hline
    \end{tabular}
    \endgroup
        \begin{tablenotes}
        \item[1] \scriptsize {
        `$*$' means that the condition has noise in the holder pose.
        }
    \end{tablenotes}
    \vspace{-4mm}
\end{table*}
\begin{figure}[t]
    \centering
    \includegraphics[width=0.99\columnwidth]{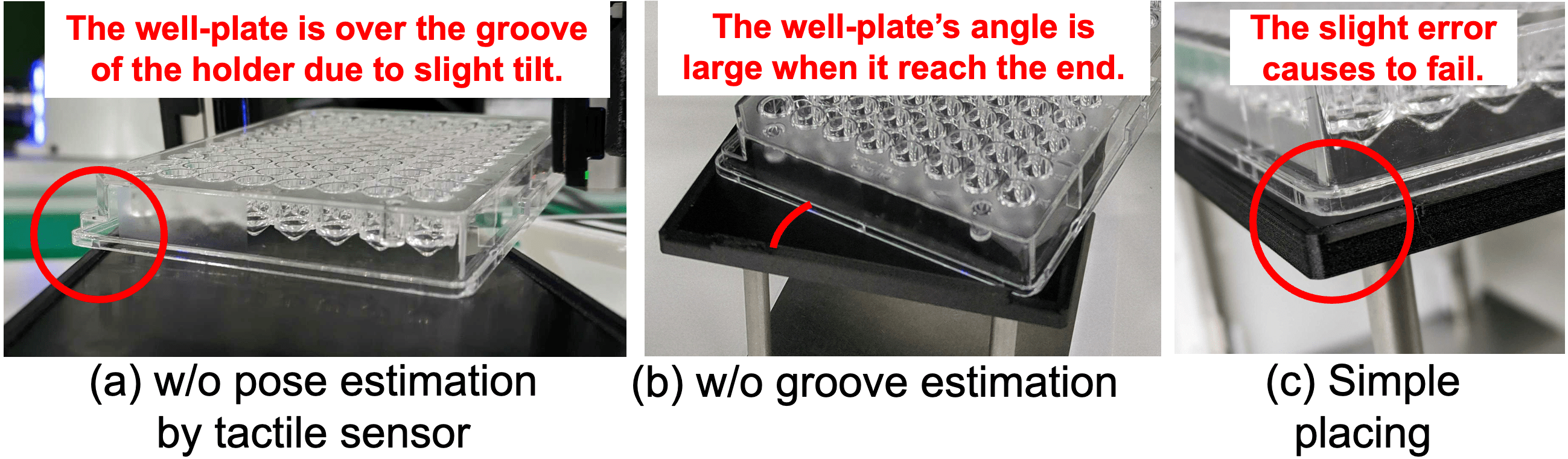}
    \caption{Major failure modes in comparison method.}
    \label{fig:result_peg_insertion}
\end{figure}

\subsection{Results of the Well-plate Placing}
\label{sec:result_peg_insertion}
As an evaluation experiment, we let the robot grasp the well-plate placed on the holder~(A) and place it onto the holder~(B).
Table~\ref{tab:result_peg_insertion} shows the success number of the well-plate placement on the holder~(B) and holder~(B)'s translation and rotation displacement measured by arUco during the placing process.
We defined placement success as the well-plate being properly aligned within the groove of the holder and failure as any misalignment, such as the well-plate being placed on the raised groove or falling outside of it.
We evaluated the performance under the following six conditions:
\begin{enumerate}
    \item Two scenarios regarding the calculation of fingertip position: one where it is adjusted based on the displacement of the adaptive finger and another where it remains constant regardless of the adaptive finger displacement.
    \item Two scenarios in regard to well-plate pose manipulation: one where pose estimation using the tactile sensor described in \cref{sec:method_pose_estimation} is used, and another where pose estimation is not executed. Without pose estimation, Fig.~\ref{fig:method}~(6) is not performed and the process proceeds from Fig.~\ref{fig:method}~(5) to Fig.~\ref{fig:method}~(7). 
    \item Two scenarios during the placing process: one where the raised groove's direction estimation described in \cref{sec:method_peg_insertion} is utilized, and another where the raised groove's direction estimation is not executed and only admittance control is applied.
    \item Two scenarios regarding the adaptive finger: one where the adaptive finger described in \cref{sec:custom_gripper} is used, and another where the function of the adaptive finger in Fig.~\ref{fig:robot_setup} is fixed.
    \item A baseline scenario in which holders~(A) and (B) are recognized using an arUco marker, and the well-plate is simply picked up and placed (see Fig.~\ref{fig:simple_placing}).
    \item A scenario in which noise is added to the detection result of the arUco marker for holder~(B), resulting in an inaccurate pose of the placement.
    Specifically, noise $(s_{1}, s_{2}, s_{3}, 0, 0, r_{1})$ is added to the arUco marker's 6D pose (x, y, z, roll, pitch, yaw) in world coordinates.
    Translation direction $s$ is randomly $\pm3~\mathrm{\,mm}$, and rotational direction $r$ is randomly $\pm1.5$ degrees.
    Since the well-plate and holder are assumed to be level on the desk apparatus, the roll and pitch noise are set to 0 in robot coordinates.
\end{enumerate}

First, we examine the effect of condition (1).
When comparing (a) and (b) in Table~\ref{tab:result_peg_insertion}, both succeed with 100\% accuracy, independent of the presence of noise in the condition (6) for (a$^{*}$) and (b$^{*}$).
Note that `$*$' means the condition has noise in the holder pose.
From this, it can be said that the information on the fingertip pose in the measurement of adaptive finger displacement has no influence on the results.
We believe the reason for this is that in the results of \cref{sec:results_pose_estimation}, when the information on the amount of displacement of the adaptive finger was not used, the error of roll became large when estimating the pose of the well-plate.
The error becomes significant when only one of the two adaptive fingers is displaced by an external force, such as contact with the holder.
Given that the task assumes the well-plate is placed evenly on a horizontal desk, it prevents the displacement of one of the adaptive fingers.

Next, we examine condition (2).
The result of method (c) shows that the well-plate placing was never successful without pose estimation using the tactile sensor.
Even a slight tilt in the roll during the well-plate placing caused the well-plate to extend over the raised groove of the holder~(B) (Fig.~\ref{fig:result_peg_insertion}~(a)).

Next, we examine condition (3).
The results of method (d) show that without estimating the direction of the holder's groove, the placing was never successful.
Despite adjustments to the gain $k_F$, the well-plate still failed to be placed successfully.
This was because the well-plate pose was controlled by admittance control to make the well-plate parallel with holder~(B), but it reached the end of the holder~(B) before achieving parallel alignment (Fig.~\ref{fig:result_peg_insertion}~(b)).

Next, we examine condition (4).
When the adaptive finger was not used in method (e), the GelSight broke, making the experiment impossible to continue.
A compliant mechanism like the adaptive finger, as described in \cref{sec:custom_gripper}, plays a critical role in absorbing excess external forces and thus preventing damage to the robot or objects it handles.
Therefore, without the use of the adaptive finger, external force was applied directly to the GelSight, leading to its breakage.
As this is an unavoidable and recurring problem, we consider these experiments a failure and the adaptive finger a necessity.

Next, we examine condition (5).
The simple placing approach can only be successful if the pose of the holder is detected accurately.
We used an arUco marker to detect the pose of the holder.
Despite using a low-pass filter to improve accuracy, the success rate was only 60\%.
As shown in Fig.~\ref{fig:result_peg_insertion}~(c), in unsuccessful attempts, the well-plate was placed on the raised groove of the holder~(B) with a misalignment of around 1~$\mathrm{\,mm}$.
This small deviation can be problematic given that the clearance between the well-plate and the holder is less than 1~$\mathrm{\,mm}$.
Under the simple placing conditions, the success rate remained the same whether the adaptive finger was used or not.
However, we encountered a major issue of significant damage to the GelSight.

Finally, we examine condition (6).
We conducted additional experiments on successful placement methods (a), (b), and (f) from placings (a)-(g), where there was no damage to GelSight.
In these experiments, the noise was introduced to the recognition of the arUco marker in the holder~(B).
Our proposed methods (a$^{*}$) and (b$^{*}$) demonstrated a 100\% success rate, highlighting their robustness against noise in the recognition process.
On the other hand, (f$^{*}$) was never successful because a slight misrecognition of the arUco marker was fatal for the simple placing approach.

In experiments (a), (b), (a$^{*}$), and (b$^{*}$), where we used the proposed method, the holder moved about 1~$\mathrm{\,cm}$ in translation and about 5 degrees in rotation.
We believe that the risk of breaking or toppling other objects by this movement is considered to be low unless the devices are placed close enough to touch each other in a chemical or biological experiment.

In conclusion, by using our proposed methods of pose estimation and groove's direction estimation, we achieved highly accurate placement even when there were recognition errors in the holder poses.
Additionally, the amount of movement from the holder during placement was minimal, reducing the chance of disturbing the surrounding environment.
Furthermore, by using the adaptive finger, we were able to reduce the risk of damage to the GelSight.
\section{Conclusion}
\label{sec:conclusion}
In this paper, we developed a method for high precision placing of a well-plate onto a holder for laboratory automation.
Our method consists of two components.
1) The well-plate pose estimation using tactile sensors to ensure well-plate contact on the holder's groove, and 2) placing utilizing contact during sliding with the holder's groove and estimating its orientation to realize sub-millimeter order accuracy and to prevent displacement of the holder.
The sub-millimeter order placing task was realized with high accuracy with small displacement, even under noisy conditions.

\section*{ACKNOWLEDGMENT}\small
The authors thank Dr. Shin-ichi Maeda and Prof. Tadahiro Taniguchi for the many discussions about this research.
\bibliographystyle{IEEEtran} 
\bibliography{IEEEabrv,bibliography}

\begin{thebibliography}{10}
\providecommand{\url}[1]{#1}
\csname url@rmstyle\endcsname
\providecommand{\newblock}{\relax}
\providecommand{\bibinfo}[2]{#2}
\providecommand\BIBentrySTDinterwordspacing{\spaceskip=0pt\relax}
\providecommand\BIBentryALTinterwordstretchfactor{4}
\providecommand\BIBentryALTinterwordspacing{\spaceskip=\fontdimen2\font plus
\BIBentryALTinterwordstretchfactor\fontdimen3\font minus
  \fontdimen4\font\relax}
\providecommand\BIBforeignlanguage[2]{{%
\expandafter\ifx\csname l@#1\endcsname\relax
\typeout{** WARNING: IEEEtran.bst: No hyphenation pattern has been}%
\typeout{** loaded for the language `#1'. Using the pattern for}%
\typeout{** the default language instead.}%
\else
\language=\csname l@#1\endcsname
\fi
#2}}

\bibitem{fleischer2016application}
H.~Fleischer, \emph{et~al.}, ``Application of a dual-arm robot in complex
  sample preparation and measurement processes,'' \emph{Journal of laboratory
  automation}, vol.~21, no.~5, pp. 671--681, 2016.

\bibitem{yachie2017robotic}
N.~Yachie and T.~Natsume, ``Robotic crowd biology with maholo labdroids,''
  \emph{Nature biotechnology}, vol.~35, no.~4, pp. 310--312, 2017.

\bibitem{burger2020mobile}
B.~Burger, \emph{et~al.}, ``A mobile robotic chemist,'' \emph{Nature}, vol.
  583, no. 7815, pp. 237--241, 2020.

\bibitem{lim2020development}
J.~X.-Y. Lim, \emph{et~al.}, ``Development of a robotic system for automatic
  organic chemistry synthesis,'' \emph{IEEE Transactions on Automation Science
  and Engineering}, vol.~18, no.~4, pp. 2185--2190, 2020.

\bibitem{shiri2021automated}
P.~Shiri, \emph{et~al.}, ``Automated solubility screening platform using
  computer vision,'' \emph{Iscience}, vol.~24, no.~3, 2021.

\bibitem{kanda2022robotic}
G.~N. Kanda, \emph{et~al.}, ``Robotic search for optimal cell culture in
  regenerative medicine,'' \emph{Elife}, vol.~11, p. e77007, 2022.

\bibitem{garrido2014automatic}
S.~Garrido-Jurado, \emph{et~al.}, ``Automatic generation and detection of
  highly reliable fiducial markers under occlusion,'' \emph{Pattern
  Recognition}, vol.~47, no.~6, pp. 2280--2292, 2014.

\bibitem{morgan2023towards}
A.~S. Morgan, \emph{et~al.}, ``Towards generalized robot assembly through
  compliance-enabled contact formations,'' in \emph{2023 IEEE International
  Conference on Robotics and Automation (ICRA)}.\hskip 1em plus 0.5em minus
  0.4em\relax IEEE, 2023, pp. 8010--8016.

\bibitem{okumura2022tactile}
R.~Okumura, \emph{et~al.}, ``Tactile-sensitive newtonianvae for high-accuracy
  industrial connector insertion,'' in \emph{2022 IEEE/RSJ International
  Conference on Intelligent Robots and Systems (IROS)}.\hskip 1em plus 0.5em
  minus 0.4em\relax IEEE, 2022, pp. 4625--4631.

\bibitem{bauza2022tac2pose}
M.~Bauza, \emph{et~al.}, ``Tac2pose: Tactile object pose estimation from the
  first touch,'' \emph{arXiv preprint arXiv:2204.11701}, 2022.

\bibitem{ota2023tactile}
K.~Ota, \emph{et~al.}, ``Tactile pose feedback for closed-loop manipulation
  tasks,'' in \emph{Robotics: Science and Systems workshop}, 2023.

\bibitem{sun2022onepose}
J.~Sun, \emph{et~al.}, ``{Onepose: One-shot object pose estimation without cad
  models},'' in \emph{Proceedings of the IEEE/CVF Conference on Computer Vision
  and Pattern Recognition}, 2022, pp. 6825--6834.

\bibitem{Li2023vox}
B.~Li, \emph{et~al.}, ``{VoxDet: Voxel Learning for Novel Instance
  Detection},'' in \emph{Proceedings of the Advances in Neural Information
  Processing Systems (NeurIPS)}, 2023.

\bibitem{von2020contact}
F.~Von~Drigalski, \emph{et~al.}, ``Contact-based in-hand pose estimation using
  bayesian state estimation and particle filtering,'' in \emph{2020 IEEE
  International Conference on Robotics and Automation (ICRA)}.\hskip 1em plus
  0.5em minus 0.4em\relax IEEE, 2020, pp. 7294--7299.

\bibitem{pankert2023learning}
J.~Pankert and M.~Hutter, ``{Learning Contact-Based State Estimation for
  Assembly Tasks},'' in \emph{2023 IEEE/RSJ International Conference on
  Intelligent Robots and Systems (IROS)}.\hskip 1em plus 0.5em minus
  0.4em\relax IEEE, 2023, pp. 5087--5094.

\bibitem{anzai2020deep}
T.~Anzai and K.~Takahashi, ``Deep gated multi-modal learning: In-hand object
  pose changes estimation using tactile and image data,'' in \emph{2020
  IEEE/RSJ International Conference on Intelligent Robots and Systems
  (IROS)}.\hskip 1em plus 0.5em minus 0.4em\relax IEEE, 2020, pp. 9361--9368.

\bibitem{lach2023placing}
L.~Lach, \emph{et~al.}, ``{Placing by Touching: An empirical study on the
  importance of tactile sensing for precise object placing},'' in \emph{2023
  IEEE/RSJ International Conference on Intelligent Robots and Systems
  (IROS)}.\hskip 1em plus 0.5em minus 0.4em\relax IEEE, 2023, pp. 8964--8971.

\bibitem{wang2019robotic}
S.~Wang, \emph{et~al.}, ``A robotic peg-in-hole assembly strategy based on
  variable compliance center,'' \emph{IEEE Access}, vol.~7, pp.
  167\,534--167\,546, 2019.

\bibitem{von2020compact}
F.~von Drigalski, \emph{et~al.}, ``A compact, cable-driven, activatable soft
  wrist with six degrees of freedom for assembly tasks,'' in \emph{2020
  IEEE/RSJ International Conference on Intelligent Robots and Systems
  (IROS)}.\hskip 1em plus 0.5em minus 0.4em\relax IEEE, 2020, pp. 8752--8757.

\bibitem{fei2003assembly}
Y.~Fei and X.~Zhao, ``An assembly process modeling and analysis for robotic
  multiple peg-in-hole,'' \emph{Journal of Intelligent and Robotic Systems},
  vol.~36, pp. 175--189, 2003.

\bibitem{kim2022active}
S.~Kim and A.~Rodriguez, ``Active extrinsic contact sensing: Application to
  general peg-in-hole insertion,'' in \emph{2022 International Conference on
  Robotics and Automation (ICRA)}.\hskip 1em plus 0.5em minus 0.4em\relax IEEE,
  2022, pp. 10\,241--10\,247.

\bibitem{kalaitzakis2020experimental}
M.~Kalaitzakis, \emph{et~al.}, ``Experimental comparison of fiducial markers
  for pose estimation,'' in \emph{2020 International Conference on Unmanned
  Aircraft Systems (ICUAS)}.\hskip 1em plus 0.5em minus 0.4em\relax IEEE, 2020,
  pp. 781--789.

\bibitem{coleman2014reducing}
D.~Coleman, \emph{et~al.}, ``Reducing the barrier to entry of complex robotic
  software: a moveit! case study,'' \emph{arXiv preprint arXiv:1404.3785},
  2014.

\bibitem{gelsight2024}
``Gelsight mini,'' \url{https://www.gelsight.com/gelsightmini/}, 2024.

\bibitem{wang2021wedge}
S.~Wang, \emph{et~al.}, ``Gelsight wedge: Measuring high-resolution 3d contact
  geometry with a compact robot finger,'' in \emph{2021 IEEE International
  Conference on Robotics and Automation (ICRA)}.\hskip 1em plus 0.5em minus
  0.4em\relax IEEE, 2021.

\end{thebibliography}
\end{document}